\documentclass[runningheads]{llncs}

\usepackage{esvect}
\usepackage[T1]{fontenc}
\usepackage{amsmath,amssymb,amsfonts}
\usepackage{algorithmic}
\usepackage{textcomp}
\usepackage{xcolor}
\usepackage{url}
\usepackage{todonotes}
\usepackage{float}
\usepackage{subcaption}
\usepackage{graphicx}
\usepackage{hyperref}
\begin{document}
\bibliographystyle{plain}

\title{Self-supervised Feature Extraction for Enhanced Ball Detection on Soccer Robots}
\titlerunning{Enhanced Ball Detection on Soccer Robots}
%
\author{Can Lin\inst{1}\and
Daniele Affinita\inst{1} \and
Marco E. P. Zimmatore\inst{1}\and
Daniele Nardi\inst{1} \and
Domenico D. Bloisi\inst{2} \and
Vincenzo Suriani\inst{3}}

\authorrunning{Lin et al.}
%
\institute{Sapienza University of Rome, Rome RM 00181, Italy \email{surname@diag.uniroma1.it}\and International
University of Rome UNINT, Rome RM 00147, Italy
\email{domenico.bloisi@unint.eu}\and
University of Basilicata, Potenza PZ 85100, Italy
\email{vincenzo.suriani@unibas.it}}
\maketitle              
\begin{abstract}

Robust and accurate ball detection is a critical component for autonomous humanoid soccer robots, particularly in dynamic and challenging environments such as RoboCup outdoor fields. 

However, traditional supervised approaches require extensive manual annotation, which is costly and time‑intensive. 
To overcome this problem, we present a self-supervised learning framework for domain-adaptive feature extraction to enhance ball detection performance. The proposed approach leverages a general-purpose pretrained model to generate pseudo-labels, which are then used in a suite of self-supervised pretext tasks —including \emph{colorization}, \emph{edge detection}, and \emph{triplet loss}— to learn robust visual features without relying on manual annotations. Additionally, a model-agnostic meta-learning (MAML) strategy is incorporated to ensure rapid adaptation to new deployment scenarios with minimal supervision. A new dataset comprising 10,000 labeled images from outdoor RoboCup SPL matches is introduced, used to validate the method, and made available to the community. 
Experimental results demonstrate that the proposed pipeline outperforms baseline models in terms of accuracy, F1 score, and IoU, while also exhibiting faster convergence. 

\end{abstract}

\section{\textbf{Introduction}}
RoboCup is one of the World’s largest robotics competitions, with more than 2,000 participants and 300 teams from 45 different Countries in 2024.

RoboCup's goal is to create a team of autonomous humanoid robots capable of defeating the FIFA World Cup human champions in 2050. 
Recent advancements in humanoid robotics have been far beyond what could have been expected only a few years ago.
Now, humanoid robots can saunter down a boardwalk, making their way across uneven terrain and catching up with a human jogger.

\begin{figure}[t]
    \centering
    \includegraphics[width=0.58\linewidth]{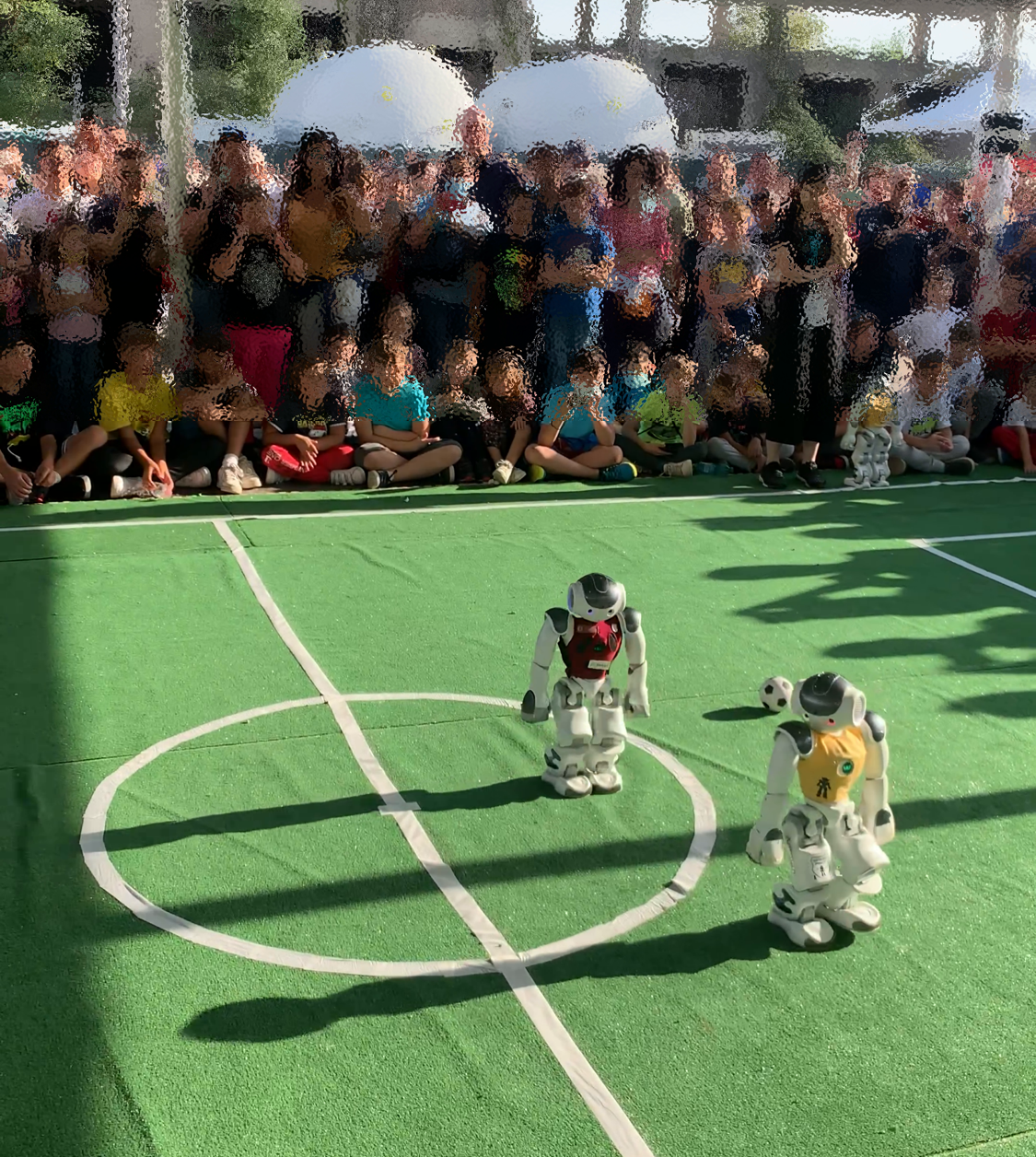}
    \caption{An example of visual challenge: Strong and long shadows on the field can confuse the perception system of the robots.}
    \label{fig:challenges}
\end{figure}

Although these improvements bode well, meeting the 2050 challenge remains extremely difficult at present.
Among the critical tasks in the RoboCup 2050 challenge, robust ball detection is essential for effective gameplay.
For example, dealing with illumination changes is still an issue for soccer robots playing outdoors (see Fig. \ref{fig:challenges}).

In RoboCup Standard Platform League, the introduction of the Realistic Ball has pushed the team to adopt deep learning approaches to detect the ball, despite the hardware constraints of the current platform, the NAO Robot \cite{albani2016deep}. 

One of the main requirements of Deep Learning is the availability of labeled data. Over the years, many datasets have been released, encompassing various tasks. However, these datasets tend to be very general, aiming to cover a broad domain. This is convenient for training general-purpose neural networks. However, when deploying models in specific applications, such generality can become a limitation. Models trained on broad datasets often struggle to adapt to narrower domains, especially if their capacity is limited \cite{zhang2024low}. In particular, general-purpose neural networks often fail to handle challenging or rare examples within a specific task, leading to reduced performance in critical scenarios \cite{hendrycks2021natural}. 
Therefore, in the case of a specific application, it is preferable to tailor the network to the target domain. This, however, requires a labeled dataset for the specific task, which is often not available online due to the niche nature of the application. As a result, data must be collected and annotated manually, a process that is both expensive and time-consuming.

To mitigate this, it would be appealing to reuse general-purpose neural networks available in pretrained form. Although they may not solve the target task directly, their embedded knowledge can be used in the early stages of the pipeline. This opens the door to self-supervised learning strategies that reduce the need for costly manual annotations while still adapting the model effectively to the specific application.

In this work, we propose a self-supervised learning pipeline that leverages a pretrained general-purpose model to generate approximate labels for a domain-specific task. These pseudo-labels are then used in a series of pretext tasks designed to extract meaningful features, which in turn enhance the performance of a downstream ball classification task. By combining the general knowledge encoded in the pretrained model with targeted self-supervised learning and a final fine-tuning on the target task, we aim to improve ball detection in complex and dynamic soccer environments, contributing to the broader goal of autonomy in RoboCup robotics.

The main contributions of this work are:
\begin{enumerate} 
\item A self-supervised learning framework that leverages a pretrained general-purpose model to generate approximate labels, which are then used to train a specialized ball detection model in complex soccer environments. 
\item A RoboCup-domain application of a loss function that accelerates convergence during the training process, enhancing the efficiency of the self-supervised learning pipeline. 
\item A dataset of 10,000 labeled images from outdoor soccer matches, collected and annotated specifically for this task, to further improve model performance. Few examples from the dataset are shown in Fig. \ref{fig:dataset}.
\end{enumerate}

The remainder of the paper is organized as follows.
Section \ref{sec:relwork} discusses related work about self-supervised learning approaches in Computer Vision.
Section \ref{sec:methodology} presents our approach based on the extraction of deep features to enhance the ability of the model using a self-supervised approach and reducing the dependence on labeled data.
Section \ref{sec:results} shows quantitative experimental results on publicly available data that demonstrate the effectiveness of the proposed approach.
Finally, conclusions and future directions are drawn in Section \ref{sec:conclusions}.

\section{\textbf{Related Work}}
\label{sec:relwork}

Self-supervised learning (SSL) has emerged as a powerful paradigm in robotics and computer vision, particularly for tasks that require learning representations from unlabeled data. SSL methods, including contrastive learning and pretext tasks, have been widely studied for their ability to improve feature extraction and representation learning without relying on large labeled datasets \cite{SSlSurvey,bootstrap,SSLtwins}. 

Recent studies have shown that SSL is particularly beneficial for detection tasks, as models trained with self-supervised objectives can learn meaningful representations that generalize well across different domains \cite{9462394}. However, one limitation of SSL is its tendency to overlook prior knowledge embedded in existing models. Unlike supervised learning approaches that leverage extensive labeled datasets, SSL methods often rely on implicit feature learning, which may not fully exploit structured priors available in pre-trained networks \cite{SSlSurvey}.

A prevalent approach in SSL involves devising pretext tasks for networks to solve, where the networks are trained by optimizing the objective functions associated with these tasks. Pretext tasks show two key characteristics. First, deep learning techniques are used to extract features that help solve pretext tasks. Second, instead of relying on external labels, supervision is derived directly from the data, a process known as self-supervision. 
The most commonly used approaches fall into four main categories: context-based methods, contrastive learning (CL), generative models, and contrastive generative approaches \cite{SSlSurvey}. 

Context-based methods exploit inherent data relationships like spatial arrangement. Generative algorithms such as Masked Image Modeling \cite{maskedimage} that reconstruct missing data portions. Contrastive generative methods aim to combine the benefits of both contrastive and generative approaches. Contrastive learning has been successfully applied in computer vision to learn discriminative features by pulling similar representations closer while pushing dissimilar ones apart. Approaches such as SimCLR \cite{chen2020simple} and MoCo \cite{he2020momentum} have demonstrated the effectiveness of contrastive learning in visual tasks, including object recognition and segmentation.

In some recent works, auto-labeled datasets generated by other pre-trained models have been employed to reduce manual labeling efforts \cite{specchi}. However, these pseudo-labeling approaches may introduce noisy or incorrect annotations, as the predictions of the models used for auto-labeling cannot be fully trusted. This uncertainty can propagate through the training process, potentially limiting the performance of downstream models if not properly mitigated.

\section{Methodology}
\label{sec:methodology}
\subsection{Metalearning}
Meta‑learning \cite{metalearning}, or “learning to learn”,  is used to further improve rapid adaptation to new field conditions with minimal labeled data. We adopt a generic Model‐Agnostic Meta‑Learning (MAML \cite{maml}) strategy:
\begin{enumerate}
  \item \textbf{Meta‑training:} sample small “tasks” representing different environments (e.g.\ varying illumination, camera angles). For each task, fine‑tune the shared encoder on a few pseudo‐labeled examples and then update the initialization to minimize post‑adaptation loss across tasks.
  \item \textbf{Meta‑adaptation:} when presented with a novel scenario, start from the learned initialization and perform only a handful of gradient steps on a small labeled set to obtain a task‐specific detector.
\end{enumerate}
This two‐loop process yields an encoder initialization that is both broadly preparatory (via SSL) and quickly adaptable (via meta‑learning), reducing the need for extensive manual annotation in new deployment settings.

\subsection{Proposed Approach}
In many machine learning scenarios, a target task \( T_1 \) can benefit from leveraging the knowledge acquired from a pre-trained model on a related broader task \( T_2 \). This paradigm follows a self-supervised learning (SSL) approach, operating under the assumption that \( T_1 \subset T_2 \). This assumption implies that the information necessary to solve \( T_1 \) is inherently embedded within \( T_2 \), making it possible to transfer learned representations effectively.
To elucidate this relationship, \( T_1 \subset T_2 \) means that the solution to \( T_2 \) produces outputs or intermediate representations that contain features relevant to solving \( T_1 \). This hierarchical structure allows for a staged learning approach, where a pre-trained model on \( T_2 \) serves as an initial feature extractor, refining and structuring the data before applying supervised learning to \( T_1 \).
The methodology consists of several key phases:

\begin{enumerate}
    \item \textbf{Pretrained Model for Feature Extraction:}  
    Initially, a model previously trained on \( T_2 \) is used to process the entire dataset. This model outputs structured representations that capture the essential high-level abstractions of the input data. By doing so, it transforms raw data into a robust feature space which provides a solid foundation for further analysis and learning. The extracted features summarize complex patterns and interactions within the data, thereby establishing a fundamental understanding that is crucial for subsequent tasks.

    \item \textbf{Definition of Pretext Tasks for Self-Supervised Learning:}  
    The high-level representations generated by the pretrained model are leveraged to formulate pretext tasks, which drive the learning process in a self-supervised manner. These auxiliary tasks are specifically designed to enable the model to learn from the inherent structure of the data without relying on manual labeling. By defining the appropriate pretext tasks, the representation space is structured in a way that is more closely aligned with the objectives of \( T_1 \). This intermediary step not only reinforces the transfer of learned knowledge from \( T_2 \) but also cultivates a latent space that is primed for the specific challenges of \( T_1 \).

    \item \textbf{Task-Specific Fine-Tuning:}  
    Following the establishment of a meaningful representation space through pretext tasks, the model undergoes a task-specific fine-tuning phase tailored for \( T_1 \). During this phase, the model is refined using supervised learning techniques, where ground truth labels pertinent to \( T_1 \) are leveraged to optimize the performance further. This fine-tuning process involves adjusting the weights and biases of the network so that the learned features become highly attuned to the nuances and requirements of the target task. Consequently, the model's predictive accuracy and generalization capabilities on \( T_1 \) are enhanced, ensuring that it performs optimally on the end goal.
\end{enumerate} 

\begin{figure}[t]
    \centering
    \includegraphics[width=\linewidth]{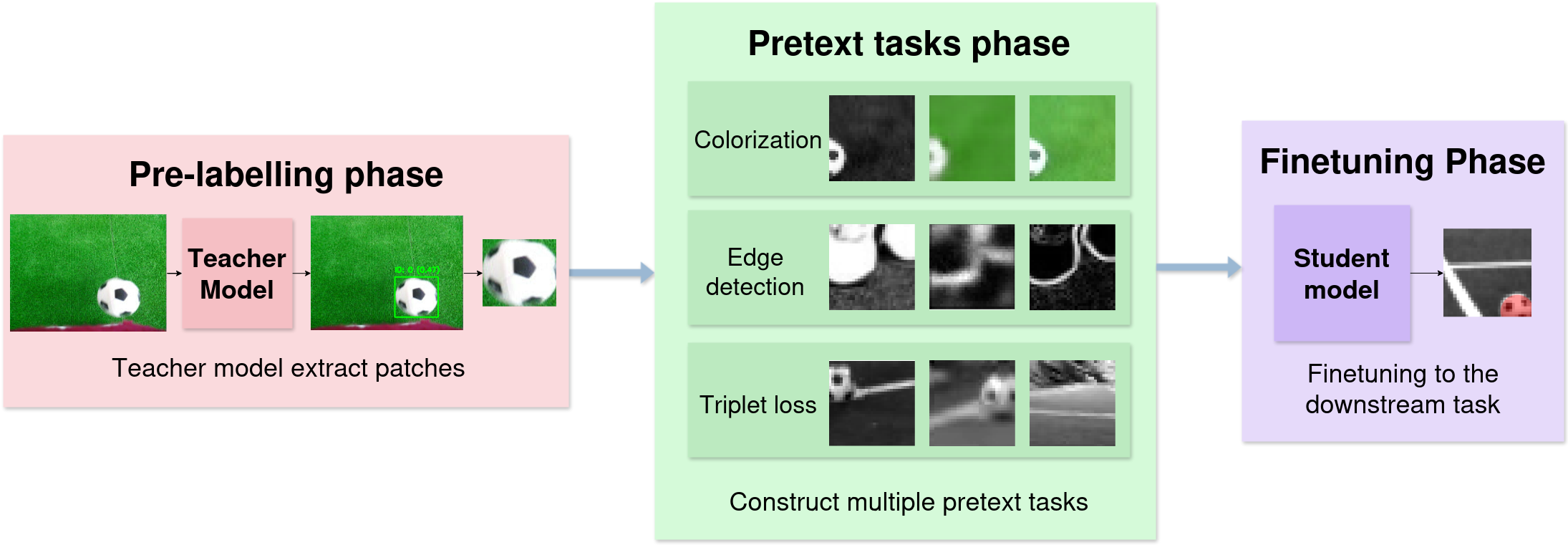}
    \caption{Overview of the data preparation pipeline: from the automatic pseudo-annotation using a pretrained YOLO‑World model, and (c) manual refinement to produce the final training, validation, and test splits.}
    \label{fig:pipeline}
\end{figure}

\section{\textbf{Experimental Setup}}
\label{sec:results}

To measure the performance of our approach, we have built and released a dataset that is made publicly available at \url{https://sites.google.com/diag.uniroma1.it/sslballperceptor}, along with extra material.

Four different metrics and five existing methods have been considered to compare the detection performance.  

\subsection{\textbf{Dataset Creation}}
\begin{figure}[t]
    \centering
    \includegraphics[width=0.75\linewidth]{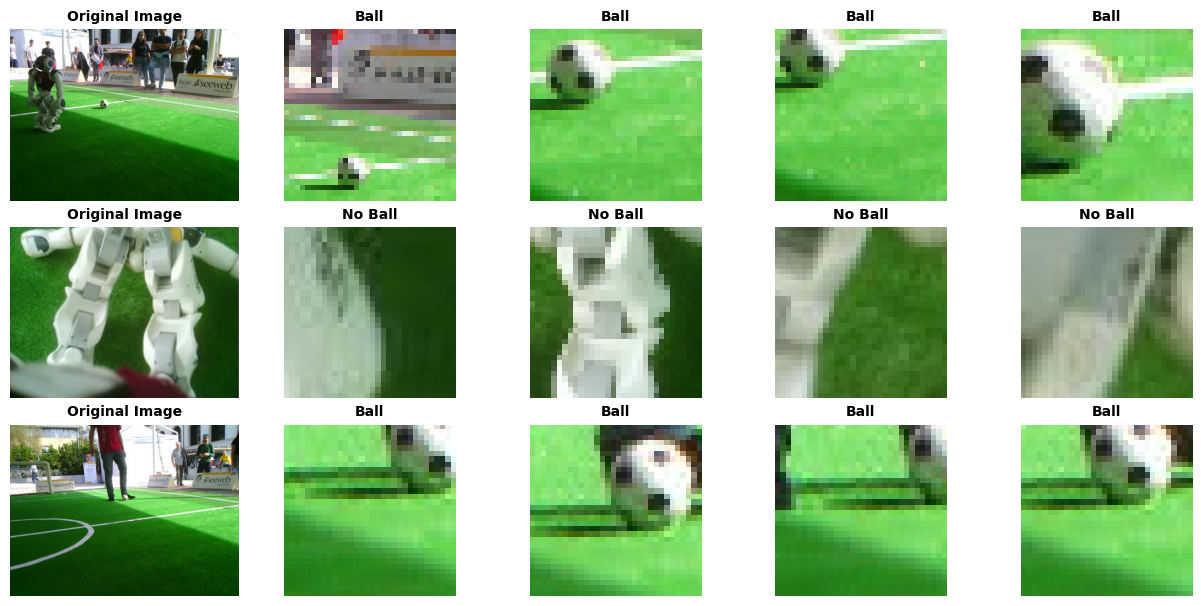}
    \caption{A few examples taken from the dataset with the original image along with the patches we created.}
    \label{fig:dataset}
\end{figure}
\subsubsection{\textbf{Retrieving logs from Nao Robots}}
We logged data from the NAO robots and extracted them using the tools provided by SimRobot \cite{simrobot}. This tool streamlined the process of accessing logs stored on multiple robots, allowing us to retrieve all the recorded images. Specifically, for each video, we extracted every 20th frame to reduce redundancy while retaining sufficient temporal diversity. 
\subsubsection{\textbf{Pre-labelling}}
Following the extraction of image frames from the NAO robot logs, we performed an initial annotation step to identify the ball and robots within each frame. For this purpose, we employed YOLO-World, a vision-language model pre-trained on a diverse set of large-scale datasets spanning object detection, grounding, and image-text tasks. The model was used to automatically generate bounding boxes around the objects of interest, serving as a preliminary labeling step. This pre-labelling process significantly accelerated dataset construction by providing annotations that could be refined in subsequent stages. By the end of the process, we had gathered approximately 14,000 images. Fig. \ref{fig:pipeline} shows the data preparation pipeline from the pre-labelling phase, while Fig. \ref{fig:dataset} shows a few examples taken from the released dataset.
\subsection{Self-Supervised Pretext Tasks and Model Architectures}
In this work, we design a suite of self-supervised pretext tasks that capture complementary aspects of the visual domain from semantic to structural information to facilitate robust feature learning without the need for labeled data. This is shown in the second phase of Fig. \ref{fig:pipeline}. 
\begin{figure}[t]
    \centering
    \includegraphics[width=0.8\linewidth]{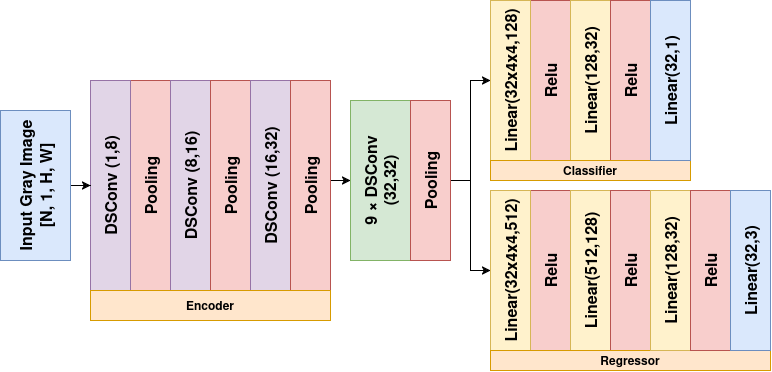}
    \caption{Base architecture consisting of an input image, a feature extraction backbone, a regressor head predicting (x, y, r), and a classification head.}
    \label{fig:base}
\end{figure}
All the tasks share a common feature extraction backbone, while task-specific decoders or heads and loss functions guide the training process. In addition, we integrate Model-Agnostic Meta-Learning (MAML) \cite{maml} to further enhance the generalization of our learned representations.

\subsubsection{Shared Feature Extraction Backbone}
All models are built upon a base architecture (Fig. \ref{fig:base}) comprising:
\begin{itemize}
    \item \textbf{Depthwise Separable Convolutions} \cite{mobilenet}: Three initial convolutional layers producing 8, 16, and 32 output channels, respectively, using \(3 \times 3\) kernels with stride 1 and padding 1 to preserve spatial resolution.
    \item \textbf{Backbone:} A stack of nine additional Depthwise Separable Convolutional layers (with some layers including residual connections) that output feature maps with 32 channels.
    \item \textbf{Max Pooling:} \(2 \times 2\) max pooling is applied after each convolutional stage to downsample spatial dimensions while retaining salient features.
\end{itemize}
Task-specific modules (such as decoders or additional heads) are appended to this backbone.

\subsubsection{Triplet Loss Task}
\begin{figure}[t]
    \centering
    \includegraphics[width=0.7\linewidth]{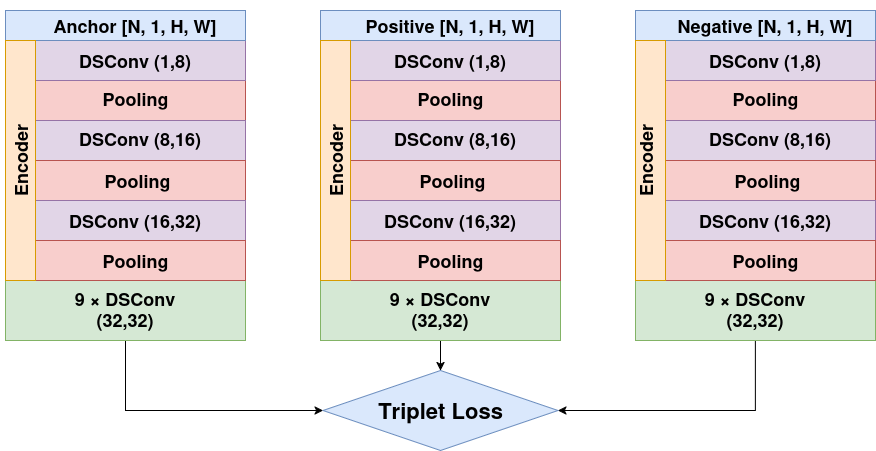}
    \caption{Triplet‑loss embedding network: the shared backbone outputs a 32×4×4 feature map, which is projected into an embedding space where anchor, positive, and negative patches are compared to enforce discriminative clustering.}
    \label{fig:triplet}
\end{figure}

The triplet loss task aims to learn a discriminative embedding space that clusters semantically similar inputs. The model (see Fig. \ref{fig:triplet}) is constructed upon the shared backbone as follows:
\begin{itemize}
    \item \textbf{Architecture:} The feature extractor comprises three initial Depthwise Separable Convolutional layers (denoted as \textit{Conv1}, \textit{Conv2}, and \textit{Conv3}) followed by the nine-layer backbone, yielding a final feature map of size \(32 \times 4 \times 4\).
    \item \textbf{Loss Function:} For an anchor sample (a patch containing the ball), a positive sample (another ball-containing patch with similar appearance), and a negative sample (a patch without the ball), the triplet loss is defined as:
    \[
    L = \max \Big( d(a, p) - d(a, n) + \alpha,\, 0 \Big)
    \]
    where \(d(a, p)\) and \(d(a, n)\) denote the distances between the anchor-positive and anchor-negative pairs, respectively, and \(\alpha\) is a predefined margin. To accelerate convergence, triplet mining \cite{hermans2017defense} is used to select the most challenging triplets, particularly cases where the negative sample is closest to the anchor in the embedding space.
\end{itemize}
\subsubsection{Colorization Task}
\begin{figure}[t]
    \centering
    \includegraphics[width=0.8\linewidth]{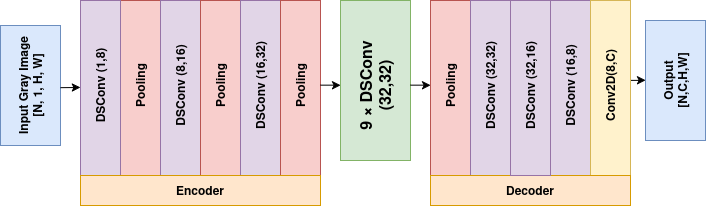}
    \caption{Both the colorization and edge detection architectures adopt an encoder-decoder structure, the only difference is the C channel in the output layer.}
    \label{fig:encoder}
\end{figure}

The colorization task challenges the model to convert grayscale images into full-color images by predicting the missing chrominance data, thereby enforcing learning of contextual and semantic scene information. Its general architecture is depicted in Fig. \ref{fig:encoder}
\begin{itemize}
    \item \textbf{Encoder:} The shared feature extractor is used.
    \item \textbf{Decoder:} The decoder consists of three transposed convolutional (deconvolutional) layers with ReLU activations, followed by a final \(2\text{D}\) convolutional layer mapping the intermediate features to three output channels corresponding to an RGB image.
    \item \textbf{Loss Function:} The model is trained to minimize the discrepancy between the predicted color image and the ground truth RGB image. In our experiments, given the limited color variability (e.g., green fields and white balls), a Mean Squared Error (MSE) loss has been shown to work sufficiently well.
\end{itemize}
\subsubsection{Edge Detection Task}
The edge detection task involves predicting edge maps from grayscale images, highlighting intensity transitions that correspond to object boundaries. It uses the same encoder-decoder architecture as the colorization task, with the only difference being the final output layer, which produces a single-channel edge map. The model is trained using Mean Squared Error (MSE) loss against edge maps generated by applying the Sobel operator to the input images.

\subsection{\textbf{Experimental Results}}
The architecture of the ball detection model remains unchanged across all the experiments to ensure fair comparison. The model, depicted in Fig. \ref{fig:base}, consists of a lightweight depthwise separable convolutional feature extractor with 13k parameters. Moreover, two fully connected heads output the class (70k parameters) and the detection (332k parameters), respectively. 
To assess the effectiveness of our self-supervised pretraining strategies, we evaluated the performance of the base model using four key metrics: \textbf{Loss}, \textbf{Accuracy}, \textbf{F1 Score}, and \textbf{Intersection over Union (IoU)}. Each model was trained on the SPQR team's dataset with weights initialized from different pretext tasks. As a baseline, we used the same architecture trained from random weight initialization without any pretraining.

\begin{table}[t]
\centering
\caption{Comparison on different metrics for the considered pretext task weights}
\begin{tabular}{|l|c|c|c|c|}
\hline
\textbf{Pretext Task Weights} & \textbf{Loss ($\downarrow$)} & \textbf{Accuracy ($\uparrow$)} & \textbf{F1 Score ($\uparrow$)} & \textbf{IoU ($\uparrow$)} \\ \hline
MAML                          & \textbf{1.72119}       & 0.96129                  & \textbf{0.90398}                 & 0.78397            \\ \hline     
Edge Detection                & 1.75486       & \textbf{0.96142}                  & 0.90298                 & 0.80340            \\ \hline
Triplet Loss                  & 1.76519       & 0.96129                  & 0.90257                 & 0.79018            \\ \hline
Colorization (MSE)            & 1.77139       & 0.95869                  & 0.89667                 & \textbf{0.80776}            \\ \hline
Base Model (Random Init)      & 1.77378       & 0.95675                  & 0.89001                 & 0.79621            \\ \hline
\end{tabular}
\label{tab:results}
\end{table}

\begin{table}[t]
\centering
\caption{Comparison of mAP in the IoU range 0.70–0.95}
\begin{tabular}{|l|c|}
\hline
\textbf{Model}              & \textbf{mAP$_{0.70\text{–}0.95}$ ($\uparrow$)} \\ \hline
MAML              & \textbf{0.41502}                            \\ \hline
Base Model                     & 0.38501                                    \\ \hline
\end{tabular}
\label{tab:map70-95}
\end{table}

\begin{figure}[t]
    \centering
    \includegraphics[width=0.7\linewidth]{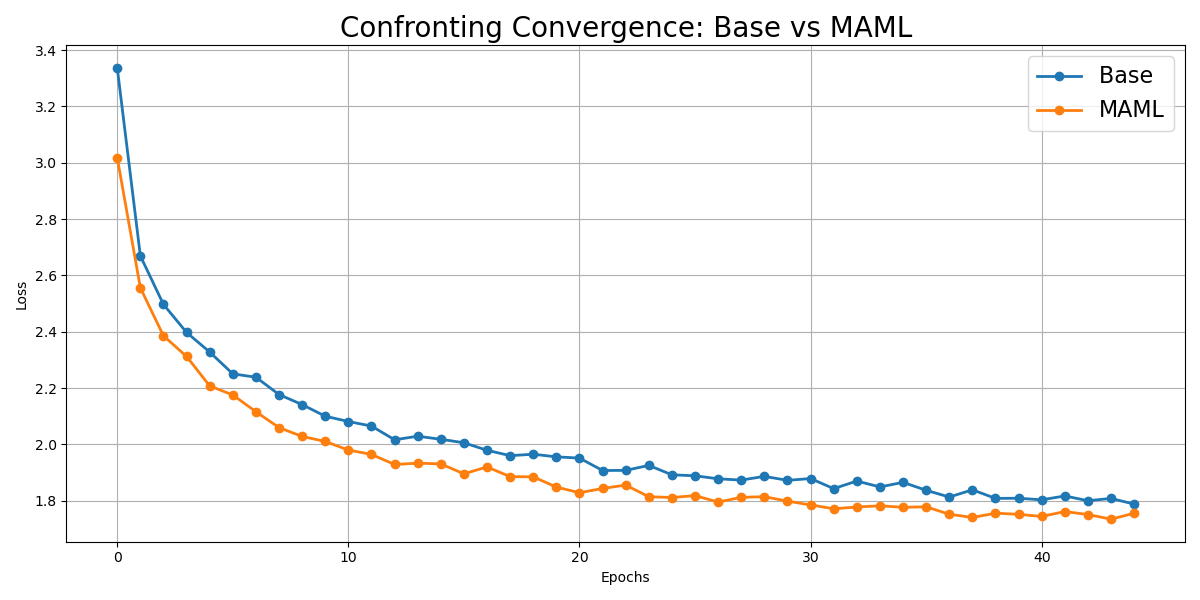}
    \caption{Validation loss curves comparing random initialization (“Base Model”) and MAML‑initialized training: the MAML‑pretrained model converges faster and attains a lower overall loss within the first few epochs.}
    \label{fig:comparison}
\end{figure}

As shown in Table~\ref{tab:results}, initializing the base model with pretext task weights consistently leads to improved performance across all evaluation metrics compared to training from scratch. Among these, the model pretrained using \textbf{Model-Agnostic Meta-Learning (MAML)} demonstrated particularly strong results.
In addition to its strong quantitative performance, the MAML-initialized model exhibited significantly faster convergence during training. While other models and the baseline required more epochs to reach optimal results, the MAML-based model consistently stabilized within the first few epochs, as depicted in Fig. \ref{fig:comparison}. 

Importantly, this rapid convergence suggests that the model may require fewer labeled samples to achieve comparable or superior performance—a critical advantage in robotic settings where data labeling is both time-consuming and resource-intensive. By reducing reliance on annotated data while also shortening training time, MAML offers a highly efficient pretraining strategy. These findings highlight its practical value for real-world deployment in data-constrained environments.

When we tighten the evaluation to the IoU range 0.70–0.95, the gap between our method and the best baseline widens: the baseline achieves a mAP of 0.3850, whereas our proposed pipeline reaches 0.4150, as shown in Table \ref{tab:map70-95}. This $\approx$ 0.03 absolute boost (nearly 8\% relative) under higher‑IoU criteria shows that our self‑supervised and meta‑learning approach lifts average performance but also delivers strong gains on the challenging cases where precise localization matters most (e.g., small, partially occluded, or low‑contrast balls). In other words, the features we learn are more robust in challenging scenarios, yielding even better improvements when evaluated under more stringent detection requirements.

\section{Conclusions}
\label{sec:conclusions}

In this paper, we introduced a self‑supervised learning framework to enhance ball detection on autonomous soccer robots operating in challenging outdoor environments. By leveraging a general‑purpose pretrained model to generate pseudo‑labels, we defined three complementary pretext tasks—triplet embedding, colorization, and edge detection—to learn domain‑specific visual features without manual annotation. 
We further incorporated a Model‑Agnostic Meta Learning (MAML) strategy to obtain an initialization that adapts rapidly to new lighting conditions and field variations with only a few labeled samples.

Our experiments on a newly collected dataset of outdoor RoboCup SPL images demonstrate that all self‑supervised initializations outperform a model trained from scratch in terms of training loss, accuracy, and F1 score. Notably, the MAML‑based initialization achieved the lowest loss (1.72) and highest F1 score (0.904), while the colorization and edge‑detection tasks yielded the best Intersection‑over‑Union improvements. Moreover, the MAML‑initialized model converged within just a few epochs, suggesting a reduced need for extensive annotation and shorter training times—critical advantages for real‑world robotic deployments.
These findings validate the effectiveness of combining pseudo‑label generation, multiple self‑supervised objectives, and meta‑learning to tackle domain adaptation in robotic vision. 

As future work, we plan to extend our pipeline to diverse outdoor scenarios, not related to the soccer domains, and investigate automated pseudo‑label refinement to further boost performance.

\section*{Acknowledgments}
We acknowledge partial financial support from PNRR MUR project PE0000013-FAIR.

\bibliography{references}

\end{document}